\documentclass[10pt, a4paper]{article}

\usepackage[]{lrec-coling2024} % this is the new style
\usepackage{graphicx}
\usepackage{amsmath, amssymb}
\usepackage[ruled, vlined]{algorithm2e}
\usepackage{float}
\usepackage{booktabs}
\usepackage{multirow}
\usepackage{tabularx}
\usepackage{refcount}
\usepackage{subfig}

\title{Advancing Semi-Supervised Learning for Automatic Post-Editing: Data-Synthesis by Mask-Infilling with Erroneous Terms}

\name{Wonkee Lee$^{12*}$\thanks{$^*$ Work completed while the author was a student at POSTECH.}, Seong-Hwan Heo$^2$, Jong-Hyeok Lee$^2$} 

\address{$^1$LG AI Research, $^2$POSTECH \\
         wklee@lgresearch.ai, \{hursung1, jhlee\}@postech.ac.kr}
         
\abstract{
Semi-supervised learning that leverages synthetic data for training has been widely adopted for developing automatic post-editing (APE) models due to the lack of training data.
With this aim, we focus on data-synthesis methods to create high-quality synthetic data.
Given that APE takes as input a machine-translation result that might include errors, we present a data-synthesis method by which the resulting synthetic data mimic the translation errors found in actual data.
We introduce a noising-based data-synthesis method by adapting the masked language model approach, generating a noisy text from a clean text by infilling masked tokens with erroneous tokens.
Moreover, we propose selective corpus interleaving that combines two separate synthetic datasets by taking only the advantageous samples to enhance the quality of the synthetic data further.
Experimental results show that using the synthetic data created by our approach results in significantly better APE performance than other synthetic data created by existing methods.
 \\ \newline \Keywords{Automatic post-editing, synthetic-corpus generation, masked language modeling} }

\begin{document}

\maketitleabstract

\section{Introduction}
\label{sec:intro}
\begin{figure}[t]
    \centering
    \includegraphics[width=\columnwidth]{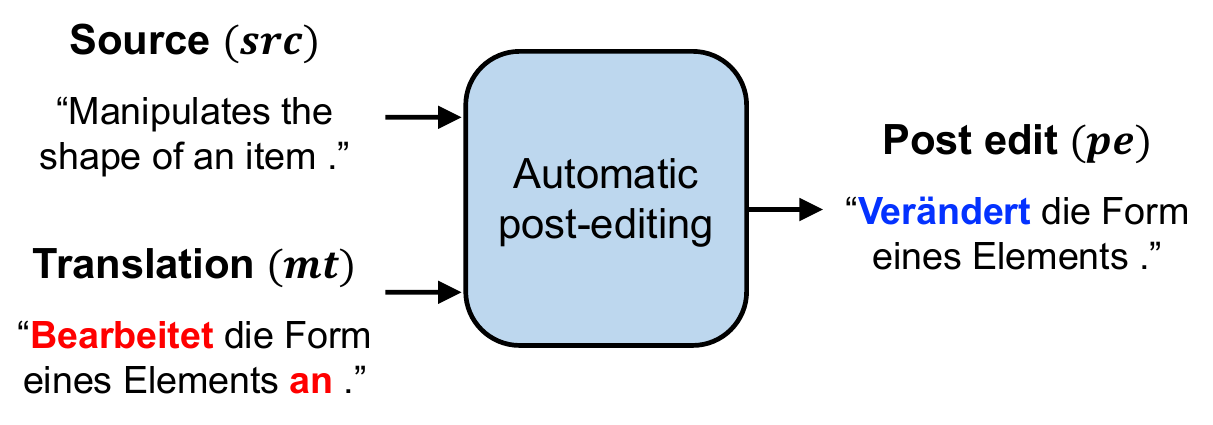}
    \caption{
        Overview of the APE process for an English-German translation.
        Bold highlights indicate erroneous words in $mt$ and post-edited words in $pe$.
    }
    \label{fig:APE}
\end{figure}

Automatic post-editing (APE)~\citep{chatterjee-etal-2015-exploring, chatterjee-2018:findings} is a study of correcting errors in machine translation (MT) outputs to provide high-quality (publishable) translations.
APE can be conceptually framed as a multi-source sequence-to-sequence problem, as depicted in Figure~\ref{fig:APE}, which concurrently takes both a source text ($src$) and its machine-translated text ($mt$) to generate a post-edited text ($pe$).
This process can be succinctly represented as a mapping of \mbox{$(src, mt) \to pe$}, and is typically addressed through supervised learning methods.

Accordingly, APE models necessitate the use of triplet data \mbox{$(src, mt, pe)$}, commonly referred to as an \textbf{APE triplet}, for supervised learning.
This data is established on the underlying assumption \citep{bojar-etal-2015-findings} that $pe$ serves as the \textbf{minimum correction} of $mt$, and both $src$ and $pe$ are assumed to be \textbf{free from error} and \textbf{semantically equivalent}.
Despite the potential of APE models, the scarcity of high-quality human-made (gold) APE data poses a significant challenge in the development of robust APE models.
To overcome this problem, semi-supervised learning, which leverages synthetic data in addition to gold data for training, has been instrumental in building APE models.

The central focus of this approach lies in the data-synthesis method, which aims to generate synthetic data that meets both quality and quantity requirements. Specifically, the resulting synthetic data should possess sufficient volume and accurately reflect the inherent characteristics found in the gold data, such as statistical properties.
For this purpose, numerous studies \citep{junczys-dowmunt-2016:log-linear_combinations, negri-2018:escape, lee-etal-2020-noising, lee-etal-2021-adaptation} have proposed methods to generate synthetic APE data, including the methods \citep{negri-2018:escape, lee-etal-2020-noising, lee-etal-2021-adaptation} that produce synthetic APE data using a bilingual parallel corpus, which have received significant attention (detailed in Section \ref{sec2}).
Although these have been proven beneficial to APE training, their synthetic data have drawbacks in that (1) the included translation errors excessively outnumber those in gold data \citep{negri-2018:escape,lee-etal-2021-adaptation} or (2) the errors are not likely to appear in gold data \citep{lee-etal-2020-noising}.

In this study, we introduce a data-synthesis approach through the application of a noising mechanism to the masked language model (MLM) framework~\citep{devlin-etal-2019-bert,conneau2019cross}.
This technique, which we have termed \textbf{MLM noising}, aims to address and mitigate the shortcomings prevalent in the existing data synthesis methods.
The objective of employing this method is to ensure that the resulting synthetic data exhibit translation errors that closely replicate those found in gold data, both in quality and quantity.
Specifically, we first manipulate a clean target text (i.e., a reference text in a parallel corpus) by replacing a specific portion with \texttt{[MASK]} tokens according to the error statistics from gold data and then let our MLM substitute each \texttt{[MASK]} with an \textbf{erroneous token} that is likely to appear in gold data.
We note that filling masks with erroneous tokens deviates from the standard approach of MLM, which aims to predict the correct tokens.

Furthermore, we contend that our MLM noising method and the existing method \citep{negri-2018:escape} may offer complementary benefits due to their orthogonal advantages.
To leverage the strengths of both approaches, we introduce a corpus ensemble method \textbf{selective corpus interleaving}.
This method selectively incorporates advantageous samples from two synthetic datasets, choosing those that most closely resemble the gold data.
We experimentally demonstrated that our synthetic data significantly improves the model performance compared to other synthetic data created by existing methods.

In summary, our key contributions to this work are as follows:
\begin{itemize}
    \item \textbf{MLM noising:} We introduce MLM noising, a method that adapts the masked language model (MLM) for creating synthetic APE data. This approach strategically inserts \texttt{[MASK]} tokens into clean target texts and replaces them with unnatural tokens that are likely to appear as translation errors in gold data.
    \item \textbf{Selective Corpus Interleaving:} We present a corpus ensemble strategy designed to combine two distinct synthetic datasets. This method makes an integrated dataset by carefully selecting the most advantageous samples from two different datasets that better resemble the characteristics of gold data.
\end{itemize}

\section{Background} \label{sec2}

\subsection{Problem Statement}

APE has been mainly framed as a dual-source sequence-to-sequence learning problem $(src, mt) \to pe$, taking two inputs: $src$, which is responsible for providing contextual information to identify translation errors, and $mt$, which is the object of correction.
Formally, let $D = \left\{(\mathbf{x}, \widetilde{\mathbf{y}}, \mathbf{y} )_i\right\}_{i=1}^{n}$ denotes a set of $n$ APE triplets (whether they are gold or synthetic triplets), where $\mathbf{x}=(x_i \dots x_{T_x})$, $\widetilde{\mathbf{y}}=(\widetilde{y}_1 \dots \widetilde{y}_{T_{\widetilde{y}}})$, and $\mathbf{y}=(y_1 \dots y_{T_y})$ indicate $src$, $mt$, and $pe$, respectively\footnote{These notations will be applied across all mathematical formulations presented in this paper.}.
An APE model learns to the estimate parameters $\theta$ that maximize the conditional probability on $pe$ as:
\begin{equation}
    P(\mathbf{y}|\mathbf{x, \widetilde{y}}) \approx \mathrm{\underset{\theta}{\arg\max}} \  \sum_{i=1}^{T_y} \log P_{\theta}(y_i \mid y_{<i}, \mathbf{x},\widetilde{\mathbf{y}}).
\end{equation}

\subsection{Existing Data-Synthesis Methods} \label{sec2: existing method}
In this section, we outline existing methods that employ a \textbf{bilingual parallel corpus} to create synthetic APE triplets.
In section 2.2.1, we provide an overview of an early method that continues to be widely used. Subsequently, in sections 2.2.2 and 2.2.3, we discuss subsequent studies that have expanded this foundational approach.

\begin{figure}[t]
    \centering
    \includegraphics[width=\columnwidth]{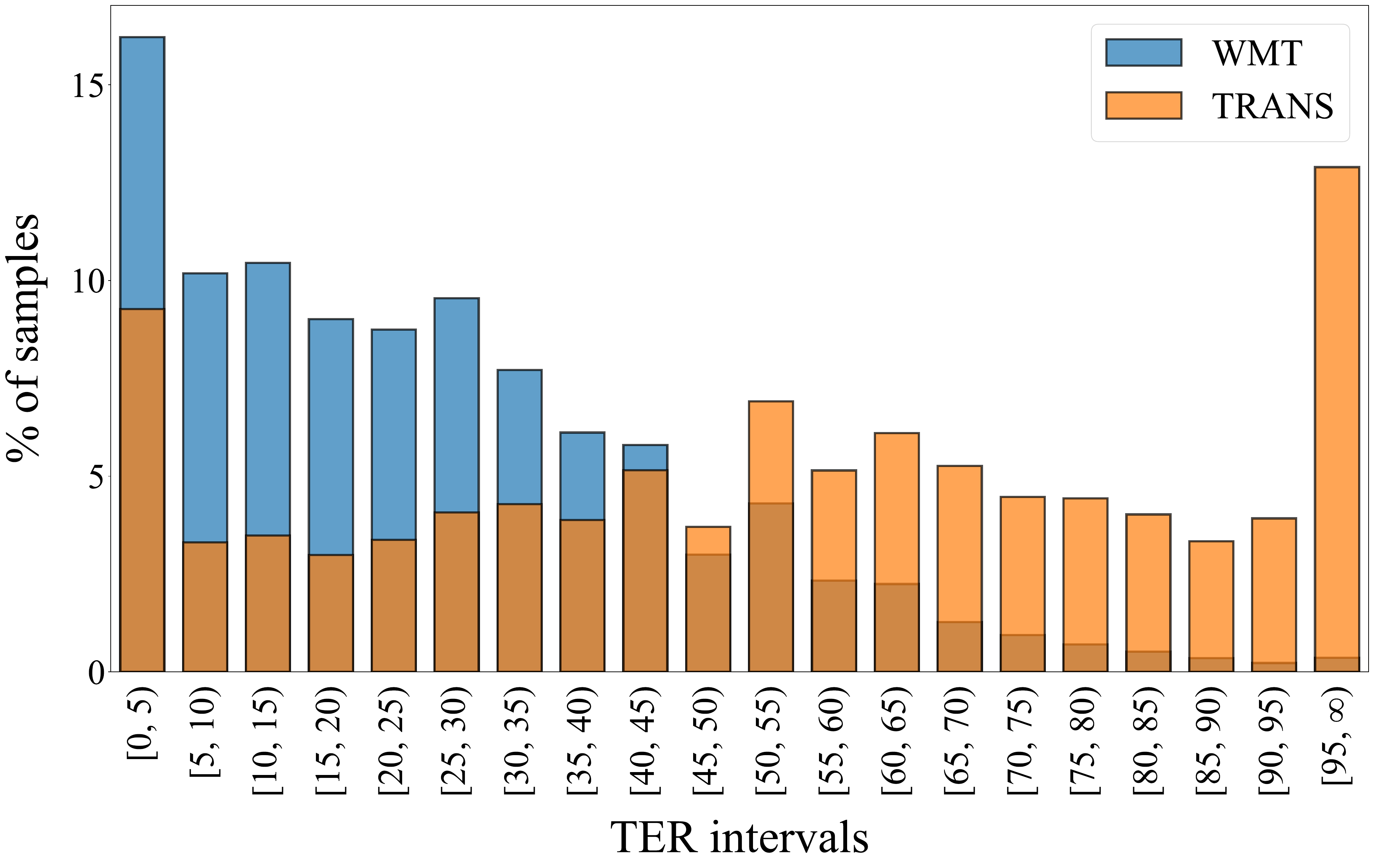}
    \caption{
    Categorical distributions of gold (WMT) and synthetic (\textsc{Trans}) APE data, representing the proportion of samples belonging to a particular interval of translation error rate (TER) [\%], a similarity measure based on the edit distance between $mt$ and its target (i.e., $pe$ or $ref$).
    }
    \label{fig:dist}
\end{figure}

\subsubsection{Translation Approach (\textsc{Trans})}\label{sec2_2: trans}
Translation approach (\textsc{Trans}), introduced in an early study by \citet{negri-2018:escape}, motivates the generation of synthetic APE data using a parallel corpus. This method remains the predominant method used in many studies.
Given that the parallel corpus consists of a source ($src$) and the corresponding reference translation ($ref$) pairs, namely, ($src, ref$), this method configures synthetic APE triplets in the form of \mbox{($src, mt, ref$)}, in which $src$ and $ref$ are taken from the parallel corpus and $mt$ is the machine translation result of $src$.

Despite its simplicity, this method has an advantage in mirroring the attributes of gold data, potentially leading to consistency in quality with gold data.
Specifically, $src$ and $ref$ in the parallel corpus are supposed to be error-free and have identical semantics, replicating the relationship observed between $src$ and $pe$ in gold APE data.
Additionally, $mt$ is generated following the same process as applied to gold data.

However, a critical limitation of this approach is the lack of assurance that the $ref$ (serving as a proxy for $pe$) represents the minimal correction required for $mt$.
This is because $ref$ is generated independently of $mt$---that is, $ref$ is not created by directly correcting $mt$---thereby deviating from the fundamental principle that gold APE data should follow.
Consequently, the synthetic data may exhibit a significantly higher number of errors compared to gold data, potentially prompting the APE model to apply overly aggressive corrections to $mt$. 
This discrepancy, as demonstrated by the differing edit-distance distributions between $mt$-$ref$ and $mt$-$pe$ (Figure~\ref{fig:dist}), indicates considerable potential for enhancing the quality of the method.
By addressing this discrepancy and more closely aligning with the characteristics of gold data, we anticipate the production of higher-quality synthetic data.
 
\subsubsection{Back-Translation Approach (\textsc{BT})}
This approach~\citep{lee-etal-2021-adaptation} modifies the \textsc{Trans} triplet ($src, mt, ref$) by replacing the machine-translated text ($mt$) with a synthesized counterpart ($\widetilde{mt}$), thus yielding the new configuration: ($src, \widetilde{mt}, ref$). 
This is accomplished through an adaptation of the back-translation (\textsc{BT}) technique, which traditionally generates synthetic source texts from target texts, by incorporating the APE process.
This method has proven effective in reducing the edit distance to $ref$ when $\widetilde{mt}$ is used instead of $mt$, thereby mitigating the problem encountered by the \textsc{Trans} method.
This method introduced two approaches for producing $\widetilde{mt}$:
\begin{itemize}
    \item Forward generation (\textsc{BT-fg}): This approach involves applying a ($src, mt$) pair to the APE process to create a partially corrected version ($\widetilde{mt}$), aiming for reduced distance to $ref$.
    \item Backward generation (\textsc{BT-bg}): This approach applies a ($src, ref$) pair to the reversed APE process, $(src, pe) \to mt$, to produce $\widetilde{mt}$ that likely contains translation errors, presenting it as a partially degraded reference.
\end{itemize}

However, the degree of reduction in the edit distance achieved by using their $\widetilde{mt}$ is insignificant, so the discrepancy remains, and moreover, the resulting synthetic data themselves improve the APE performance less than \textsc{Trans} (Section~\ref{sec:result}).

\begin{figure*}[t]
    \centering
    \includegraphics[width=\textwidth]{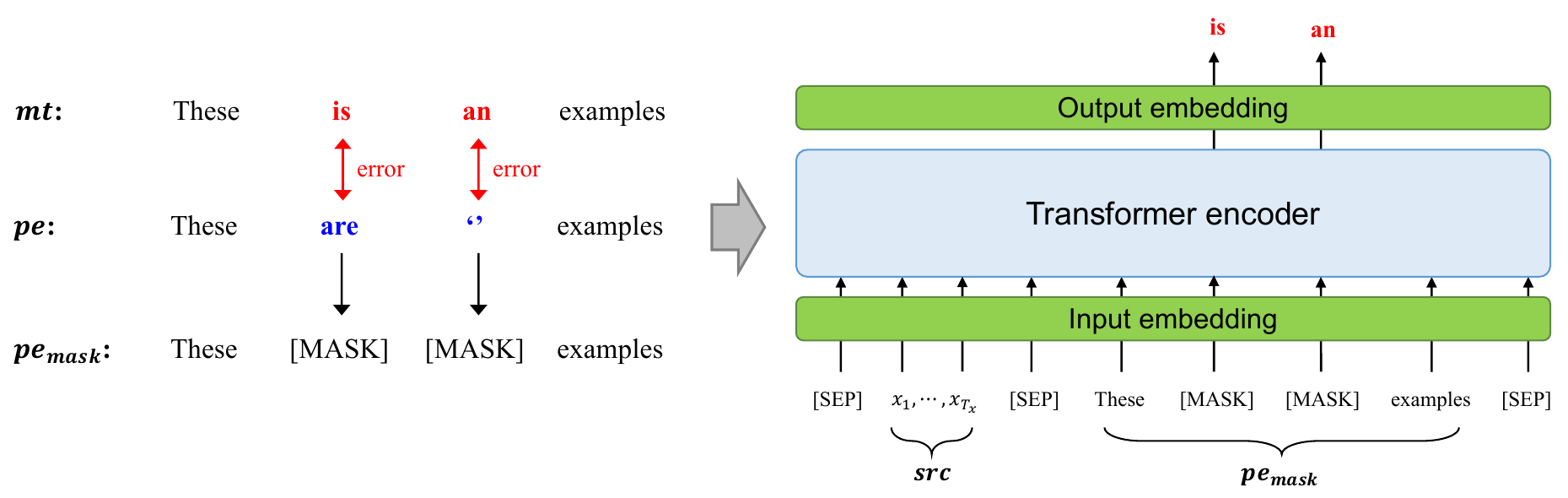}
    \caption{
    Overall architecture of our MLM noising model.
    }
    \label{fig:mlm}
\end{figure*}

\subsubsection{Random Noising Approach (\textsc{Rand})}
As in the \textsc{BT} approach, this method \citep{lee-etal-2020-noising} substitutes $mt$ of \textsc{Trans} triplet with its synthetic machine translation $\widetilde{mt}$.
This method introduces random noises (\textsc{Rand}) into $ref$ to create $\widetilde{mt}$, aligning the number of introduced noises with the error distribution observed in gold data.
Specifically, for each token in $ref$, the method probabilistically applies one of four editing operations---\textit{keep}, \textit{insert}, \textit{delete}, or \textit{substitute}---reflecting translation errors.
This selection is based on the categorical probability distribution of these operations derived from gold data, yielding a collection of synthetic APE triplets ($src, \widetilde{mt}, ref$).

The primary advantage of this approach is its capacity to produce synthetic data whose error frequency mirrors that of gold data. 
However, a significant challenge arises from the qualitative disagreement between the synthetic and gold data, caused by the stochastic nature of the noising process. 
This randomness is evident in the procedures of word selection for insertion and substitution, where choices are made indiscriminately from the dictionary, and deletions similarly affect randomly selected words. 
As a result, their synthetic machine translations $\widetilde{mt}$ may incorporate words that machine translation systems are unlikely to generate, leading to a qualitative deviation from the expected output of such systems.

\section{Approach}
\subsection{Overview and Motivation} \label{sec:overview_and_motive}
It is apparent from the existing methods that \textsc{BT-bg} and \textsc{Rand} have unique advantages.
\textsc{BT-bg} is prone to generating translation errors that are likely to be found in ($src, pe$) pairs by learning of $(src, pe) \to mt$.
On the other hand, \textsc{Rand} effectively limits the number of errors to align with the distribution found in gold data.
In response to these observations, we introduce a noising method by adapting the masked language model (MLM), termed \textbf{MLM noising}.
This method learns $(src, pe) \to mt$ and leverages the mask-infilling technique of MLM to create synthetic machine translations $\widetilde{mt}$ by introducing plausible translation errors into the reference ($ref$) of a parallel corpus while maintaining the frequency of these introduced errors to align with the error distribution found in gold data.
Consequently, it presents a hybrid method that combines the advantages of both the \textsc{BT-bg} and \textsc{Rand} methods, effectively balancing error realism with controlled error quantity.

In addition, the \textsc{Trans} method is conspicuous in that it utilizes raw machine translations ($mt$) that are not synthesized, thus closely replicating the generation process of $mt$ found in gold data if the error frequency in $mt$ aligns with that of gold data.
To capitalize on the distinct advantages of both the \textsc{Trans} and MLM noising methods, we propose a corpus ensemble method called \textbf{selective corpus interleaving}.
This approach merges two distinct synthetic datasets (e.g., the datasets created by \textsc{Trans} and MLM noising) into a single, enhanced dataset. 
This integration is achieved by comparing corresponding samples from the two datasets and selectively incorporating those that more accurately reflect the attributes of gold data, thereby expecting to improve the overall quality and effectiveness of the synthetic data.

\subsection{MLM Noising} \label{sec:mlm-noising}

\subsubsection{Architecture and Objective}
Given a corrupted text in which some tokens are replaced with \texttt{[MASK]}, the objective of MLM \citep{devlin-etal-2019-bert} is to reconstruct the original text from this corrupted text by infilling masks with original tokens using the Transformer encoder~\citep{vaswani-2017:attention_is_all_you_need}.
Inspired by studies \citep{kumar2020data,tuan-etal-2021-quality} that employed MLMs for the data augmentation, we adapt MLM to produce a synthetic $mt$ (i.e., $\widetilde{mt}$) that serves as a proxy for $mt$ in our synthetic APE triplet.

As shown in Figure~\ref{fig:mlm}, given an APE triplet ($src, mt, pe$), we mask $pe$ tokens that are aligned to \textbf{mistranslated tokens} in $mt$, after which we feed $src$ along with this masked $pe$ ($pe_{mask}$) and let the model learn to restore these \textbf{erroneous}
$mt$ tokens from the \texttt{[MASK]} tokens.
In other words, this training objective indicates that the model learns to predict translation errors that may occur in ($src, pe$) pairs, leading to the learning of $(src, pe) \to mt$ that \textsc{BT-bg} likewise learns.
We also note that our MLM noising can be regarded as a \textbf{cross-lingual MLM} \citep{conneau2019cross} because the two bilingual inputs, $src$ and $pe$, constitute a cross-lingual representation by which the prediction is made on \texttt{[MASK]}.
Formally, given that $\widehat{\mathbf{y}}$ denotes $pe_{mask}$, the objective function can be defined as 
\begin{equation}
\begin{split}
    \mathrm{\underset{\theta}{\max}} \ \   \mathcal{J}(\theta) &= P_\theta (\mathbf{\widetilde{y}} \mid \mathbf{x}, \widehat{\mathbf{y}}) \\
    &\approx \sum_i m_i \log P_\theta (\widetilde{y}_i \mid \mathbf{x}, \widehat{\mathbf{y}}) \\
    &= \sum_i m_i \log \frac{\exp \left(H_\theta(\widehat{y}_i)E^\top_\theta(\widetilde{y}_i)\right)}{\sum_{y'} \exp\left(H_\theta(\widehat{y}_i)E^\top_\theta(y')\right)},
\end{split}
\end{equation}
where $H_\theta(\cdot)$ and $E_\theta(\cdot)$ are the hidden vector from Transformer and the embedding vector, respectively; and $m_i=1$ indicates that $\widehat{y}_i$ is masked to correspond with mistranslated $\widetilde{y}_i$.

\subsubsection{Training Data Configuration}
For the learning process, it is imperative to construct a set of triplets in the form of ($src, pe_{mask}, mt$).
At first, the alignment between $mt$ and $pe$ is necessary to construct $pe_{mask}$, but it is generally invisible in data.
As an alternative, we utilize an \textbf{alignment}\footnote{We obtain the alignments by using tercom software: \url{https://github.com/jhclark/tercom.git}
} that can be obtained as a byproduct of the edit distance\footnote{The rationale for employing edit-distance alignment stems from the premise that $pe$ results from only minor correction to $mt$, with the possibility that the majority of words between them remain unchanged. Consequently, we assume that discrepancies in word alignment between $mt$ and $pe$—where aligned words are not identical—indicate that the word from $mt$ is likely an erroneous translation.} calculation between $mt$ and $pe$.
Given the edit-distance alignments from $pe$ to $mt$ denoted as $a=\{(y, \widetilde{y})_i\}_{i=1}^{T_{\widetilde{y}}}$, each alignment corresponds to one of four edit operations\footnote{Due to implementation difficulties with mask-infilling, the shift operation employed in TER is not directly considered. Nonetheless, it is addressed indirectly, as it is widely recognized that a shift can be accomplished through a sequence of a deletion followed by an addition.}: \textit{keep, insert, delete, or substitute}. 
Specifically, for each $(y, \widetilde{y}) \in a$, the keep operation, where $y = \widetilde{y}$, represents that $\widetilde{y}$ is an accurate translation. 
Conversely, an alignment where $y \ne \widetilde{y}$ indicates a mistranslation, which is further categorized into (1) an insertion error if $y$ is \texttt{NULL}, indicating the addition of unnecessary token; (2) a deletion error\footnote{In practice, we ignore alignments on deletion errors because all deletion \texttt{[MASK]} are mapped onto a single output token (i.e., null token `'), consequently causing serious deletion-biased predictions. Instead, we simulate deletion errors at the inference time.} if $\widetilde{y}$ is \texttt{NULL}, indicating the omission of necessary token; and (3) a substitution error when both $y$ and $\widetilde{y}$ are present but differ, indicating incorrect translation. 
Finally, we construct $pe_{mask}$ by replacing $pe$ tokens, which are aligned with mistranslated $mt$ tokens, with \texttt{[MASK]} tokens.

Due to the limited availability of gold data, we transform synthetic data from the \textsc{Trans} triplets ($src, mt, ref$) into ($src, ref_{mask}, \widetilde{mt}$) for our training needs. 
Notably, \textsc{Trans} typically exhibits a higher error rate compared to gold data, necessitating the use of $\widetilde{mt}$ to approximate the edit distance between $mt$ and $pe$ found in gold data.

To align the error distribution of synthetic data with that of gold data, we first calculate the proportion of masking tokens based on the error rates observed in the gold data (by sampling from the categorical distribution in Figure~\ref{fig:dist}).
For each $ref$ in the \textsc{Trans} triplet, we sample the number of tokens ($k$) to be masked and find the total tokens ($n$) that are aligned with mistranslated $mt$ tokens based on edit-distance alignment.
Next, we randomly choose $k$ tokens out of the $n$ tokens, replace these with their aligned $mt$ tokens to form $\widetilde{mt}$, and mask them to create $ref_{mask}$. 
This procedure yields a set $\mathcal{S} = \{(\widehat{y}, \widetilde{y}_i)\}_{i=1}^{|\mathcal{Y}|}$, where \(|\mathcal{Y}|={n \choose k}\) represents the total number of potential masking scenarios, each leading to distinct pairs of ($ref_{mask}, \widetilde{mt}$).
The model is trained to recognize these masking patterns across iterations, aiming at generating appropriate errors at various positions by
\begin{equation}
    \mathrm{\underset{\theta}{\max}} \ \ \mathbb{E}_{(\widehat{\mathbf{y}}, \mathbf{\widetilde{y}})\sim\mathcal{S}} \left[ \sum_i m_i \log P_\theta (\widetilde{y}_i \mid \mathbf{x}, \widehat{\mathbf{y}}) \right].
\end{equation}
We present our pseudo algorithm for constructing $ref_{mask}$ from the \textsc{Trans} triplet ($src, mt, ref$) in Algorithm \ref{algo:algo1}.

\SetKwInOut{Input}{Input}
\SetKwInOut{Output}{Output}
\newlength\mylen
\newcommand\myinput[1]{%
  \settowidth\mylen{\KwIn{}}%
  \setlength\hangindent{\mylen}%
  \hspace*{\mylen}#1\\}
\begin{algorithm}[t]
    \caption{\small{$ref_{mask}$ for training}}
    \label{alg:cap}
    \small
    \KwIn{$\mathbf{y}, \widetilde{\mathbf{y}}$ \quad \textit{\# $ref$ and $mt$}\newline
        $D$ \textit{\# distribution of error rate from gold data}  \newline
        $a= \{(y, \widetilde{y})_i\}_{i=1}^{T_{\widetilde{y}}}$ \quad \textit{\# alignments} 
    } 

    \KwOut{$\widehat{\mathbf{y}}$ }
    $\widehat{\mathbf{y}} \leftarrow \{\}$
    
    $e \leftarrow \operatorname{Edit\_distance}(\widetilde{\mathbf{y}}, \mathbf{y})$ \quad \textit{\# actual error rate}
    
    $\hat{e} \sim Categorical(D)$ \quad \textit{\# sampled error rate}
    
    $a^{err} \leftarrow \lbrace(y, \widetilde{y}) \mid \forall{a},  \widetilde{y} \ne \texttt{NULL}, y \neq \widetilde{y}  \ \rbrace$
    
    \If{$e > \hat{e}$}{
            num\_mask $\leftarrow \lceil \left(  \operatorname{Len}(\mathbf{y}) * \hat{e} \right)\rceil$ 
            
            $\hat{a} \leftarrow \operatorname{random \_ choice}(a^{err}, \text{num\_mask})$  
        }
    \Else{
    	$\hat{a} \leftarrow a^{err}$\
    }
    \For{each $ (y, \widetilde{y}) \in\ a $}{
    	    \If{$ \langle y, \widetilde{y} \rangle \in\ \hat{a} $}{
    	        $\widehat{\mathbf{y}} \leftarrow \operatorname{Append}( \widehat{\mathbf{y}}, \text{\texttt{[MASK]}})$
    	    }
    	        
    	   \Else{
    	        $\widehat{\mathbf{y}} \leftarrow \operatorname{Append}( \widehat{\mathbf{y}}, y)$
    	   }
    	}\label{algo:algo1}
\end{algorithm}

\subsubsection{Inference: Synthetic APE Data Construction}
Once the training has finished, we use our MLM model to produce $\widetilde{mt}$ from the parallel corpus, yielding a new set of synthetic APE triplets ($src, \widetilde{mt}, ref$).
For every ($src, ref$) pair, we initiate the masking process for $ref$, wherein each token is masked \textbf{stochastically}\footnote{Note that the alignment used in the training phase is intrinsically unavailable during inference.} according to the categorical probability distribution\footnote{Each $\mu$ represents the probability that the corresponding edit operation occurs. This distribution can be obtained from the edit-distance calculation: \url{https://github.com/jhclark/tercom.git}} of the edit operations $\mathbf{\mu}=\{\mu_{\text{keep}}, \mu_{\text{ins}}, \mu_{\text{del}}, \mu_{\text{sub}}\}$ obtained from gold data, facilitates diverse outcomes resulting from the informed masking decisions.
We present this stochastic masking process to obtain $ref_{mask}$ in Algorithm~\ref{algo:algo2}.
Subsequently, we provide our MLM model with ($src, ref_{mask}$) and let it perform mask infilling to produce $\widetilde{mt}$.

Eventually, $\widetilde{mt}$ can hold the number of errors similar to gold data while these errors are expected to appear convincing by learning.
Moreover, because our MLM stochastically determines masking patterns of $ref_{mask}$ at every inference time, we can obtain $\widetilde{mt}$ in various forms from a single ($src, ref$) pair, making a given resource scalable.

\begin{algorithm}[t]
    \caption{\small{$ref_{mask}$ for inference}}
    \small
    \KwIn{$\mathbf{y}, \mu= \lbrace \mu_{\text{keep}}, \mu_{\text{ins}}, \mu_{\text{del}}, \mu_{\text{sub}} \rbrace$ 
    }

    \KwOut{$\widehat{\mathbf{y}}$}
    
    $\widehat{\mathbf{y}} \leftarrow \{\}$
    
    $\text{op} \sim Categorical(\mu)$
    
    \For{each $y \in \widehat{\mathbf{y}}$}{
        \If{op is keep}{
        	        $\widehat{\mathbf{y}} \leftarrow \operatorname{Append}(\widehat{\mathbf{y}}, y)$
        	    }
        \ElseIf{op is insert}{
        	        $\widehat{\mathbf{y}} \leftarrow \operatorname{Append}(\widehat{\mathbf{y}}, y)$
        	        
        	        $\widehat{\mathbf{y}} \leftarrow \operatorname{Append}(\widehat{\mathbf{y}}, \text{\texttt{[MASK]}})$
        	   }	        
        \ElseIf{op is delete}{
        	        continue
        	   }
        \ElseIf{op is substitute}{
        	        $\widehat{\mathbf{y}} \leftarrow \operatorname{Append}(\widehat{\mathbf{y}}, \text{\texttt{[MASK]}})$
                }
    } \label{algo:algo2}
\end{algorithm}

\subsection{Selective Corpus Interleaving} \label{sec:interleaving}

Recall that although \textsc{Trans} is advantageous in using non-synthesized $mt$ (which is most likely to capture the nature of $mt$ in gold data), it has a limitation in including excessive errors on $mt$ to match $ref$.
We thus hypothesize that the benefit of synthetic $mt$ that approximates the error statistics of gold data far outweighs the benefit of raw $mt$.
However, it is intuitively reasonable that a \textsc{Trans} triplet containing $mt$ whose error quantity is already similar to that of gold data can be regarded as an ideal training sample for APE, which can be even more effective than using $\widetilde{mt}$ instead (e.g., the overlapping region in Figure~\ref{fig:dist}).

In this regard, we suggest a selective corpus interleaving approach that takes only advantageous samples between \textsc{Trans} and ours depending on whether the $mt$--$ref$ edit distance is similar to the $mt$--$pe$ edit distance, thereby constructing a single enhanced synthetic dataset.
For every ($src, ref$) pair, we select either $mt$ or $\widetilde{mt}$ by applying the three-sigma rule \citep{10.2307/2684253}:
\begin{equation}
    mt =
    \begin{cases}
    mt & \text{if } \lvert \operatorname{edit}(mt, ref)-\mu \rvert \le \lambda\sigma \\
    \widetilde{mt} & \text{otherwise,}
    \end{cases}
\end{equation}
where $\operatorname{edit}(\cdot)$ denotes the edit distance; $\mu$ and $\sigma$ are the mean and standard deviation of the $mt$--$pe$ edit distances, respectively; and $\lambda \in \left[ 1, 3 \right]$ is a hyperparameter.

\section{Experiments}
\subsection{Setup}
\paragraph{Evaluation Metric}
Following the WMT APE shared task \citep{chatterjee-2018:findings}, we adopted TER ($\downarrow$) \citep{snover-2006:a_study_of_translation_edit_rate}\footnote{\url{https://github.com/jhclark/tercom}} as our primary metric, and BLEU ($\uparrow$) \citep{koehn-2007:moses}\footnote{\url{https://github.com/moses-smt/mosesdecoder}} as the secondary metric.
We conducted all evaluations case-sensitively.

\begin{figure}[t]
    \centering
    \includegraphics[width=\columnwidth]{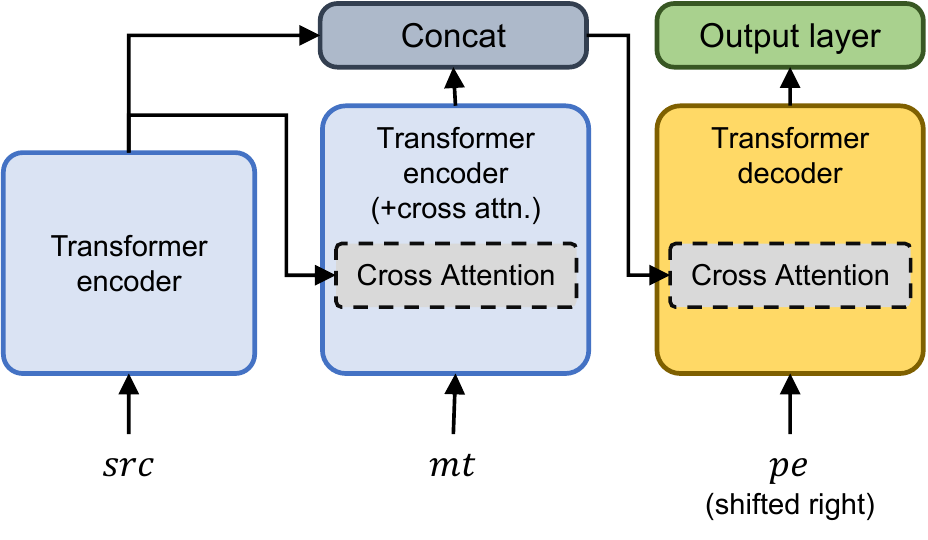}
    \caption{
        Schematic architecture of the concat-based APE model, which is among the fundamental APE models and exhibits outstanding performance.
    }
    \label{fig:ape}
\end{figure}

\begin{table*}[t]
    \centering
    \small
    \begin{tabular}{llllllllll}
        \toprule
        \multicolumn{1}{c}{\multirow{3}{*}{Approach}} & \multicolumn{2}{c}{\textbf{Test16}} & \multicolumn{2}{c}{\textbf{Test17}} & \multicolumn{2}{c}{\textbf{Test18}} & \multicolumn{2}{c}{\textbf{Test Avg.}}
        \\
        & TER{\small(\textdownarrow)} & BLEU{\small(\textuparrow)} & TER{\small(\textdownarrow)} & BLEU{\small(\textuparrow)} & TER{\small(\textdownarrow)} & BLEU{\small(\textuparrow)} & TER{\small(\textdownarrow)} & BLEU{\small(\textuparrow)} \\
        \midrule
        
        \textsc{Trans} & 16.87 & 73.95 & 17.30 & 73.08 & 17.80 & 72.41 & 17.32 & 73.15 \\
        \textsc{BT-fg} & 17.26 & 73.56 & 17.56 & 72.78 & 17.89 & 72.14 & 17.57 & 72.82 \\
        \textsc{BT-bg} & 17.61 & 73.04 & 17.60 & 72.49 & 18.01 & 71.89 & 17.74 & 72.47\\
        \textsc{Rand} & 17.23 & 73.59 & 17.61 & 72.69 & 17.81 & 72.38 & 17.55 & 72.88\\
        \midrule
        MLM Noising (w/o interleave) 
                    & 16.90 
                    & 74.03
                    
                    & 17.31
                    & 72.90
                    
                    & 17.62
                    & 72.43
                    
                    & 17.28
                    & 73.12  \\
        MLM Noising (w/ interleave) 
                    & \textbf{16.71} 
                    & \textbf{74.58} 
                    
                    & \textbf{16.74} 
                    & \textbf{73.79}
                    
                    & \textbf{17.43}
                    & \textbf{72.88}
                    
                    & \textbf{16.96}
                    & \textbf{73.75} \\
        \bottomrule
    \end{tabular}
    \caption{
        Comparison of learning effects on each synthetic data by presenting evaluation results of the model trained on each specific synthetic data.
        In each column, the best-performing result is highlighted in \textbf{bold}.
        }
    \label{comparison}
\end{table*}

\paragraph{Datasets}
We used the WMT'18 APE dataset as the gold data, a human-made English-German (EN-DE) APE dataset considered the \textit{de-facto} standard benchmark.
The WMT data consists of 23K of training data, 1K of development data, and three separate test datasets (Test16, Test17, and Test18) of 2K each.
We also used EN-DE \textsc{Trans} synthetic triplets\footnote{\url{https://ict.fbk.eu/escape/}} consisting of approximately 7M samples.
All words in the datasets we used were tokenized into subword units by SentencePiece\footnote{\url{https://github.com/google/sentencepiece}}.

\paragraph{Model Configuration}
Our MLM noising model was developed utilizing the RoBERTa architecture~\citep{liu-2019:roberta}\footnote{\url{https://huggingface.co/roberta-base}}, adhering largely to its default configuration. 
Specifically, the model comprises 12 layers, each with 12 attention heads, a hidden layer size of 768, and a feed-forward layer size of 3,072. 
For training, we employed the AdamW optimizer \citep{loshchilov2018decoupled}, configured with $\beta$ parameters of (0.9, 0.999), a learning rate of 2e-4 subject to linear decay, 7,000 warm-up steps, and managed a batch size of 384 samples.

For the implementation of an APE model, we utilized OpenNMT-py\footnote{\url{https://github.com/OpenNMT/OpenNMT-py}}, specifically deploying the ``concat-based'' model (Figure~\ref{fig:ape}) that aligns with state-of-the-art standards. 
In line with established practices, the model architecture was configured with 6 layers, 8 attention heads, hidden layer sizes of 512, and feed-forward network sizes of 2,048. The training was conducted using the Adam optimizer \citep{2015-adam} with $\beta$ set to (0.9, 0.998), following the learning rate schedule recommended by \citet{vaswani-2017:attention_is_all_you_need}, which includes 6,000 warm-up steps. The model was trained with a batch size capable of accommodating 48K tokens.

\subsection{Experimental Details}
In our experiment, we establish our baselines using four models trained on the existing synthetic datasets discussed in Section~\ref{sec2}: \textsc{Trans, BT-fg\footnote{\label{bt}\url{https://github.com/wonkeelee/APE-backtranslation.git}}, BT-bg\footnotemark[\getrefnumber{bt}],} and \textsc{Rand}.
To ensure a fair and controlled comparison, each model was trained under identical conditions, including hyperparameters, codebase, and training seed, with the sole exception of the synthetic training data. 
Furthermore, all synthetic datasets, including our own, were created using the same parallel corpus used to generate \textsc{Trans} so that they only differed in the $mt$ portion.

Since our MLM noising and \textsc{Rand} can diversify their $\widetilde{mt}$ as stochastic, we empirically used five different random seeds to construct them; namely, one out of five different $\widetilde{mt}$ was selected for each ($src, ref$) pair at every iteration during training.
The training of the MLM model lasted 3 days with 8 A100 GPUs and the APE model lasted for 12 hours with a single A5000 GPU.

\begin{table}[t]
\centering
\resizebox{\columnwidth}{!}{
    \begin{tabular}{lllll}
    \toprule
    \multicolumn{1}{c}{} & \multicolumn{2}{c}{\textbf{Test Avg.}} & \multicolumn{2}{c}{\textbf{Sample Ratio}} \\ 
    & TER{\small(\textdownarrow)} & BLEU{\small(\textuparrow)} & MLM & \textsc{Trans} \\
    \midrule
    $\lambda=0 \ (\text{MLM Noising})$ & 17.32 & 73.15 & 100.0\% & 0.0\%    \\
    $\lambda=1$ & 17.25                       & 73.26                         & 71.5\% & 28.5\%     \\
    $\lambda=2$ & \textbf{16.96}$^{\ast\ast}$ & \textbf{73.75}$^{\ast\ast}$   & 41.5\% & 58.5\%      \\
    $\lambda=3$ & 17.03$^{\ast\ast}$          & 73.48$^{\ast\ast}$            & 20.8\% & 79.2\%     \\
    $\lambda=\infty \ (\text{\textsc{Trans}})$ & 17.28 & 73.12                & 0.0\%    & 100.0\%   \\
    \bottomrule
    \end{tabular}
}
\caption{
    Effect of the selective corpus interleaving with varying $\lambda$.
    $^{\ast\ast}$ indicates the improvement is statistically significant compared to both $\lambda=0$ and $\lambda=\infty$ with $p < 0.01$.
    The \textbf{bold} highlights indicate the best result in each column.
}
\label{tab:interleaving}
\end{table}

\subsection{Results} \label{sec:result}

We evaluated the APE models, each trained on distinct synthetic datasets, by employing the WMT test datasets. This allowed us to investigate how each synthetic dataset impacts their performance.
Referring to Table~\ref{comparison}, our observations yield two key findings: (1) The utilization of synthetic data generated through MLM noising improves the APE performance compared to the baselines, and (2) augmenting this with selective corpus interleaving method further enhances the model performance. 
In light of these results, we speculate that our findings lend support to our initial hypothesis, suggesting potential benefits for APE training. Specifically, the combination of MLM noising, drawing from the strengths of \textsc{BT-bg} and \textsc{Rand}, and interleaving MLM noising with \textsc{Trans}, where the former focuses on statistics while the latter aligns more closely with gold data, demonstrates promise in improving the APE performance.

Additionally, we performed experiments employing various values of $\lambda$ for the corpus interleaving method (Table~\ref{tab:interleaving}). 
When $\lambda=0$, it signifies the exclusive use of the dataset generated by MLM noising, while $\lambda=\infty$ indicates the exclusive use of the \textsc{Trans} dataset.
Notably, our results revealed that taking approximately equal parts $(\lambda=2)$ from our MLM noising and \textsc{Trans} data led to better APE performance than the other ratios.
These findings also support our hypothesis that each synthetic dataset possesses unique advantages that complement one another.

\begin{figure}[t]
    \centering
    \includegraphics[width=0.8\columnwidth]{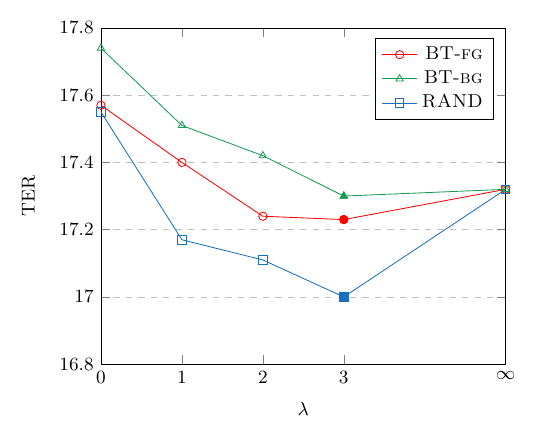}
    \caption{
        Effect of the selective corpus interleaving when applied to other existing synthetic APE data.
        The colored plots represent the best performance for each data.
    }
    \label{fig:intv}
\end{figure}

\begin{figure}[t]
    \centering
    \includegraphics[width=0.8\columnwidth]{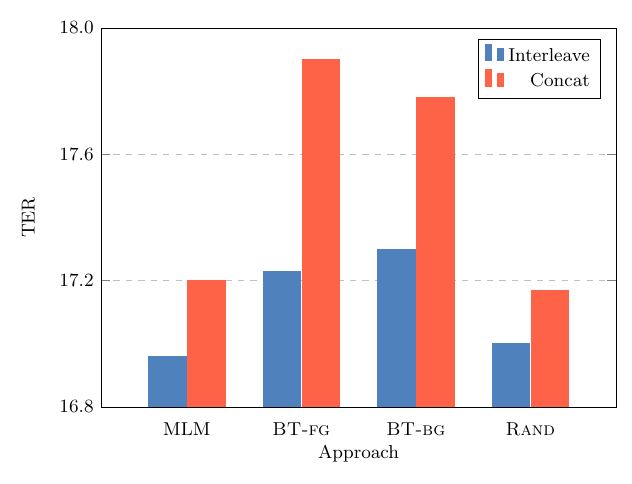}
    \caption{
        Comparison of two corpus ensemble methods applied to each synthetic APE dataset.
    }
    \label{fig:intv_concat}
\end{figure}

\section{Analysis and Discussion}

\subsection{Effect of Selective Corpus Interleaving}

\begin{table*}[t]
    \centering
    \small
    \subfloat[\label{mlm_win}][]{%
        \begin{tabularx}{.7\linewidth}{l |l}
            \toprule
            Component & Sentence \\
            \midrule
            $src$ & It is worthless without knowing where to sell it .\\
            $ref$ & Zu wissen , wo man es verkauft , ist sehr wichtig . \\ 
            \midrule
            $\widetilde{mt}$ (MLM noising) & Zu wissen , wo man es \textcolor{red}{\textbf{zu}} verkauft , \textcolor{red}{\textbf{wo sie}} sehr wichtig .\\
            $mt$ (\textsc{Trans}) & \textcolor{red}{\textbf{Es ist nichts ohne zu}} wissen , wo \textcolor{red}{\textbf{sie zu verkaufen}} .\\
            \bottomrule
        \end{tabularx}
    }
    
    \subfloat[\label{tran_win}][]{%
        \begin{tabularx}{.7\linewidth}{l |l}
            \toprule
            Component & Sentence \\
            \midrule
            $src$ & What happens if I want to leave ?\\
            $ref$ & Was geschieht , wenn ich wieder gehen will ? \\ 
            \midrule
            $\widetilde{mt}$ (MLM noising) & Was \textcolor{red}{\textbf{passiert}} geschieht , wenn ich \textcolor{red}{\textbf{ich zu}} gehen \textcolor{red}{\textbf{wollen .}}\\
            $mt$ (\textsc{Trans}) & Was \textcolor{red}{\textbf{passiert}} , wenn ich \textcolor{red}{\textbf{verlassen wollen}} ?\\
            \bottomrule
         \end{tabularx}
    }
    \caption{
        Two examples showing the comparison between $mt$ and $\widetilde{mt}$ for a given ($src, ref$) pair.
        The \textcolor{red}{\textbf{bold}} highlights indicate the mistranslated words compared to $ref$.
    }
    \label{tab:case}
\end{table*}

The versatility of the selective corpus interleaving technique allows for its application across a variety of pre-existing synthetic datasets, including \textsc{BT-fg}, \textsc{BT-bg}, and \textsc{Rand}, provided that these datasets feature synthetic machine translations $\widetilde{mt}$ aimed at reducing the edit distance to $ref$.
To investigate this utility further, we undertook additional experiments where selective corpus interleaving was employed to integrate \textsc{Trans} data with each of the aforementioned datasets (\textsc{BT-fg}, \textsc{BT-bg}, and \textsc{Rand}).
The outcomes, illustrated in Figure~\ref{fig:intv}, indicate that utilization of the selective corpus interleaving with these alternate synthetic datasets consistently improves the quality of the synthetic dataset, thereby contributing to enhanced performance of the models.

Moreover, we conducted a comparative analysis contrasting the selective corpus interleaving method with a conventional corpus ensemble technique that straightforwardly concatenates two datasets without any selective filtering.
The outcomes (Figure~\ref{fig:intv_concat}) validate that the selective corpus interleaving approach significantly surpasses the straightforward concatenation method in performance, even though the latter offers a twofold increase in data resources. 
This finding highlights the importance of excluding filtered samples, a crucial step in the effective application of the selective corpus interleaving method, underscoring its impact on improving data quality.

\subsection{Case Study}

In Table~\ref{tab:case}, we present two examples to examine the differences between our modified translation output by MLM noising, denoted as $\widetilde{mt}$, and the raw translation output, referred to as $mt$ (representing \textsc{Trans}).
The first example, shown in Table~\ref{mlm_win}, illustrates that $mt$ requires extensive post-processing to align with the reference ($ref$). In contrast, $\widetilde{mt}$, generated through our approach, necessitates fewer corrections, showing a closer resemblance to the error statistics found in gold data. 
However, in our second example, presented in Table~\ref{tran_win}, the number of errors in $mt$ is already within an acceptable range, making it more compatible with the gold standard. In such cases, the benefits of using $\widetilde{mt}$ are less pronounced, and the selective corpus interleaving method ultimately favors $mt$.

Additionally, our observations have identified several potential weaknesses in our approach:
\begin{itemize}
    \item 
    Due to the nature of the masked language modeling, all masked tokens are reconstructed independently of one another. This independence can occasionally lead to undesirable words, such as repeated words especially when consecutive positions in $ref$ are masked during the inference process. Nevertheless, this independence may be acceptable as long as it doesn't occur excessively, as our primary goal is to introduce errors intentionally.
    
    \item 
    Unlike insert and substitute operations, delete operations are applied randomly to construct the masked reference ($ref_{mask}$) for MLM noising. While some level of randomness is acceptable for our error introduction purposes, it can occasionally alter the semantics of the text (e.g., "New York City" becoming "New City").
    
    \item
    The edit-distance alignment used to create the masked post-edited text ($pe_{mask}$) for MLM training does not always correspond to semantic errors. Consequently, $\widetilde{mt}$ may sometimes contain unnatural or incorrect words.
\end{itemize}

\section {Related Work}
A similar approach involving the use of MLM for generating synthetic data was proposed by \citet{tuan-etal-2021-quality} in the context of the quality estimation task that is similar to the APE task in terms of that models are trained on pairs of ($src, mt$)
Their method shares some similarities with ours: (1) they employed MLM to create synthetic $mt$ from a parallel corpus, and (2) for a given $ref$, they randomly replaced tokens with \texttt{[MASK]} or deleted tokens and inserted \texttt{[MASK]} tokens at random positions. Subsequently, they input $src$ and the masked $ref$ into their MLM model to perform mask infilling, resulting in synthetic $mt$.

Nevertheless, there are several notable distinctions between our approach and theirs:
(1) In their masking procedure, the choice of tokens for insertion, deletion, and substitution was independent of the statistical properties of the gold data;
(2) They employed a pre-trained, off-the-shelf multilingual BERT model \citep{devlin-etal-2019-bert}, which was trained on clean text, and thus, their MLM was incapable of inducing errors in $mt$ from \texttt{[MASK]} tokens;
(3) Despite addressing multilingual texts, their multilingual BERT model did not incorporate cross-lingual features, thereby limiting its ability to establish a joint representation between $src$ and $mt$.

\section{Conclusion}

In this study, we introduce a method for adapting MLM to generate synthetic APE data from a parallel corpus. 
This method is further refined through the incorporation of selective corpus interleaving. 
We summarize our key findings as follows:
\begin{itemize}
    \item 
    We present a training approach for MLM, designed to predict translation errors from masked tokens. By utilizing this trained MLM model, we generate synthetic $mt$ data by introducing errors into $ref$ through a process known as mask infilling.
    \item
    Additionally, we propose the selective corpus interleaving method, a technique for effectively merging two distinct datasets. This method involves the selection of samples whose error profiles closely resemble those found in the gold data, bridging the gap between the datasets.
    \item
    Our experimental results and analysis highlight the importance of ensuring that synthetic data closely mirrors the error statistics of gold data.
\end{itemize}

We believe that our work offers opportunities for further extensions and improvements by incorporating different MLM techniques. For instance, approaches like ELECTRA~\citep{2003.10555} that employ adversarial learning within MLM can help determine the plausibility of mask prediction results, resulting in more natural and realistic synthetic data.

\section{Limitations}
Our method, while promising, presents several challenges that need to be addressed in the future:
\begin{itemize}
    \item 
    The edit distance-based alignments utilized in our training may not always correspond to semantic discrepancies, potentially resulting in the generation of translation errors during inference that fail to reflect actual translation inaccuracies. This discrepancy might result in the production of errors that appear unnatural.
    
    \item 
    The inherent nature of MLM, which treats each masked token independently, makes them prone to generating plausible yet contextually inappropriate tokens. Consequently, token generation from our MLM noising process may not always meet the nuanced requirements of natural language generation.
    
    \item
    Although the MLM model's error distribution aims to mimic that of gold APE data (i.e., benchmark datasets), it is crucial to recognize that these gold datasets do not capture every possible translation error. As a result, the MLM noising approach may be inherently biased towards the types of translation errors prevalent in the benchmark datasets used for training, potentially restricting the diversity of errors it generates.
\end{itemize}

\nocite{*}
\section{Bibliographical References}\label{sec:reference}

\bibliographystyle{lrec-coling2024-natbib}
\bibliography{lrec-coling2024-example}
\end{document}